%% file: manuscript.tex
\documentclass[runningheads]{llncs}
\usepackage{graphicx}
\usepackage[utf8]{inputenc} 
\usepackage[T1]{fontenc}    
\usepackage{xr-hyper}
\usepackage{hyperref}       
\usepackage{url}            
\usepackage{booktabs}       
\usepackage{amsfonts}       
\usepackage{amsmath}       
\usepackage{nicefrac}       
\usepackage{microtype}      
\usepackage{epsfig}
\usepackage{array,multirow}
\usepackage{multicol}
\usepackage{xspace}
\usepackage{color}
\usepackage{cleveref}[capitalise]

\makeatletter
\DeclareRobustCommand\onedot{\futurelet\@let@token\@onedot}
\def\@onedot{\ifx\@let@token.\else.\null\fi\xspace}

\def\eg{\emph{e.g}\onedot}

\def\etal{\emph{et al}\onedot}
\makeatother

\graphicspath{{./fig/}}

\begin{document}
\title{Self-Supervised Single-Image Deconvolution with Siamese Neural Networks}
\titlerunning{Self-Supervised Deconvolution}
%
\author{
Mikhail Papkov\inst{1} \and
Kaupo Palo\inst{2} \and
Leopold Parts\inst{1,3}
}
\authorrunning{M. Papkov et al.}
%
\institute{
Institute of Computer Science, University of Tartu, Estonia \and
Revvity Inc., Tallinn, Estonia \and
Wellcome Sanger Institute, Hinxton, United Kingdom\\
\email{mikhail.papkov@ut.ee}
}
%


\maketitle              
\begin{abstract}
Inverse problems in image reconstruction are fundamentally complicated by unknown noise properties. Classical iterative deconvolution approaches amplify noise and require careful parameter selection for an optimal trade-off between sharpness and grain. Deep learning methods allow for flexible parametrization of the noise and learning its properties directly from the data. Recently, self-supervised blind-spot neural networks were successfully adopted for image deconvolution by including a known point-spread function in the end-to-end training. However, their practical application has been limited to 2D images in the biomedical domain because it implies large kernels that are poorly optimized. We tackle this problem with Fast Fourier Transform convolutions that provide training speed-up in 3D microscopy deconvolution tasks. Further, we propose to adopt a Siamese invariance loss for deconvolution and empirically identify its optimal position in the neural network between blind-spot and full image branches. The experimental results show that our improved framework outperforms the previous state-of-the-art deconvolution methods with a known point spread function.

\end{abstract}
\section{Introduction and related work}
\label{sec: intro}

Quality enhancement is central for microscopy imaging~\cite{weigert2018content,zhang2019poisson}. Its two most important steps are noise removal and blur reduction that improve performance in downstream tasks for both humans and algorithms. 

Denoising, approaches to which were developed decades ago~\cite{dabov2007image}, was given a boost by recent advances in deep learning methods~\cite{batson2019noise2self,buchholz2019cryo,krull2019noise2void,krull2019probabilistic,lehtinen2018noise2noise,weigert2018content,lemarchand2019opendenoising,papkov2021noise2stack,moran2019noisier2noise}, which substantially outperform classical algorithms~\cite{dabov2007image,buades2011non}. A direct approach to train a denoising neural network is to provide it with pairs of low and high-quality images~\cite{weigert2018content}, but such data are rarely available in practice, which raises the need for alternative techniques. Lehtinen~\etal~\cite{lehtinen2018noise2noise} proposed to supervise the training with an independently acquired noisy copy of the input, while Batson and Royer~\cite{batson2019noise2self} concurrently with Krull~\etal~\cite{krull2019noise2void} developed a theory of a $J$-invariant blind-spot denoising network that operates on a single image in a self-supervised manner. The latter theory is based on the idea that a network cannot learn the noise that is conditionally pixel-wise independent given the signal, so learning to predict the values of masked pixels can improve the performance in restoration. Xie~\etal~\cite{xie2020noise2same} further questioned the necessity of blind spots for efficient denoising. They proposed Noise2Same with Siamese invariance loss between outputs of masked and unmasked inputs to prevent the network from learning an identity function. This advance pushed the performance of self-supervised denoising at a cost of doubled training time.

In addition to noise, equipment imperfection imposes blur on the images via a point spread function (PSF)~\cite{sage2017deconvolutionlab2}. The blur can be efficiently removed with classical iterative approaches such as Lucy-Richardson~\cite{richardson1972bayesian}, but these algorithms tend to amplify the noise with each iteration and require careful regularization and stopping criteria~\cite{laasmaa20103d}. As supervised methods are not applicable to deconvolution problem due to the ground truth inaccessibility,  Lim~\etal~\cite{lim2020cyclegan} proposed CycleGAN~\cite{zhu2017unpaired} with linear blur kernel, which performed well in both simulation and real microscopy deconvolution. However, the models of this class are notoriously hard to train~\cite{mescheder2018training} and tends to converge to a perceptually conceiving solution that does not necessarily reflect the true underlying structure~\cite{ledig2017photo}. Besides, it requires clean data examples, albeit unpaired. Deep self-supervised denoising systems were also adopted for deconvolution purposes. Kobayashi~\etal~\cite{kobayashi2020image} assumed an intermediate output of the network to be a deconvolved representation and proposed to train Noise2Self~\cite{batson2019noise2self} as a pseudo-inverse model with a known PSF kernel. This method was only applied to 2D data.

Our contribution is three-fold. Firstly, we adopt the Siamese invariance loss~\cite{xie2020noise2same} to the deconvolution task and identify the optimal neural network's outputs to apply it. Secondly, we train the first, to our knowledge, 3D self-supervised deconvolutional network without the adversarial component. Lastly, we propose to alleviate the computational costs of training a Siamese network by using the Fast Fourier Transform (FFT) for convolution with the PSF kernel.

\section{Methods}
\label{sec: methods}

\begin{figure}[h]
\begin{center}
\includegraphics[width=0.99\linewidth]{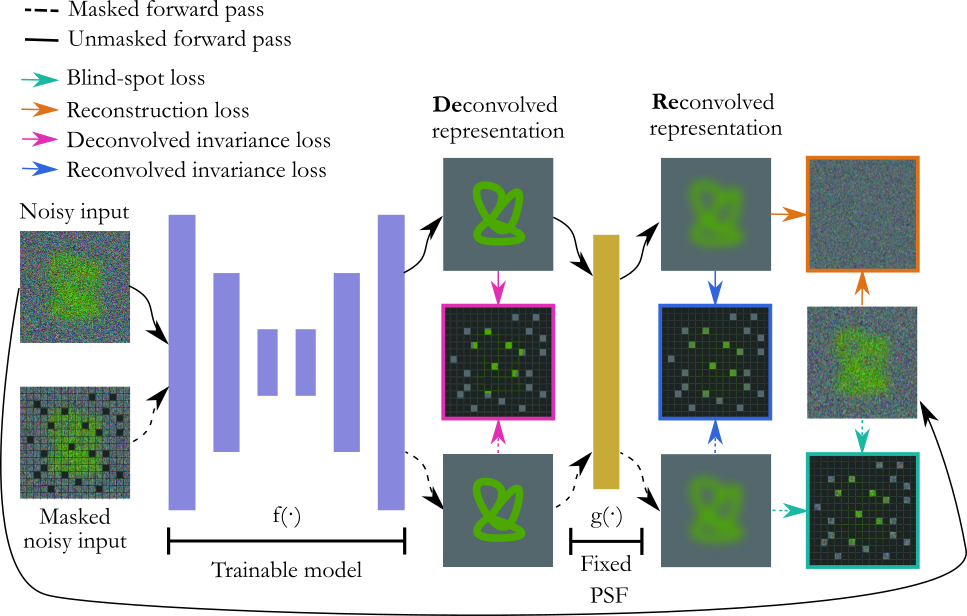}
\end{center}
   \caption{The self-supervised deconvolution framework. The trainable model $f(\cdot)$ takes noisy input $\mathbf{x}$ in unmasked forward pass (solid) and masked noisy input $\mathbf{x}_{J^c}$ in masked forward pass (dashed). The model produces two intermediate outputs (deconvolved representations). These outputs are convolved with a fixed PSF convolution $g(\cdot)$ to obtain final outputs (reconvolved representations). The deconvolved and reconvolved invariance losses are computed between the respective outputs before and after the PSF. The reconstruction loss is computed between the noisy input and the unmasked forward pass output. The blind-spot loss is computed between the noisy input and masked forward pass output, it does not require an unmasked forward pass.}
\label{fig:title}
\end{figure}

We propose a generalized self-supervised deconvolution framework. We follow Kobayashi~\etal~\cite{kobayashi2020image} and define the framework as a composition of trainable model $f(\cdot)$ followed by the fixed PSF convolution $g(\cdot)$. The framework (Figure~\ref{fig:title}) allows for both one-pass blind-spot training\cite{krull2019noise2void,batson2019noise2self,kobayashi2020image} and two-pass training~\cite{xie2020noise2same}. 

For the training, we use two inputs, unmasked $\mathbf{x}$ and masked $\mathbf{x}_{J^c}$ where the pixels at locations $J$ are replaced with Gaussian noise. We optimize the trainable model using the composition loss $\mathcal{L}(f)$ of six terms in Eq.~\eqref{eq:loss}.

\noindent\textbf{Blind-spot loss.} $\mathcal{L}_{bsp}$ is a mean-squared error (MSE) between the network's unmasked input $\mathbf{x}$ and output from the masked forward pass $g(f(\mathbf{x}_{J^c}))_J$ measured at the locations of the masked pixels $J$~\cite{krull2019noise2void,batson2019noise2self}. 

\noindent\textbf{Reconstruction loss.} $\mathcal{L}_{rec}$ is an MSE between the network's unmasked input $\mathbf{x}$ and output from the unmasked forward pass $g(f(\mathbf{x}))$~\cite{xie2020noise2same}.

\noindent\textbf{Reconvolved invariance loss.} $\mathcal{L}_{inv}$ is an MSE between the network's output from the masked forward pass $g(f(\mathbf{x}_{J^c}))_J$ and output from the unmasked forward pass $g(f(\mathbf{x}))_J$. 

\noindent\textbf{Deconvolved invariance loss.} $\mathcal{L}_{inv\,(d)}$ is an MSE between the network's output from the masked forward pass $f(\mathbf{x}_{J^c})_J$ and output from the unmasked forward pass $f(\mathbf{x})_J$ before PSF convolution $g(\cdot)$. 
Both Siamese invariance losses are computed between the network's outputs from altered (masked and unmasked) inputs at the location of the masked pixels $J$. They prevent the network from learning an identity function from $\mathcal{L}_{rec}$ minimization, which is especially important in the single-image deconvolution task.  

\noindent\textbf{Boundary losses}. $\mathcal{L}_{bound\,(d)}$ and $\mathcal{L}_{bound}$ regularize the outputs to be within $[min, max]$ boundaries before or after the PSF convolution respectively~\cite{kobayashi2020image}. These losses are measured after destandardization. We use $min = 0$ and $max = 1$.

In this work, we compare three practical cases. The first one, Eq.~\eqref{eq:bsp}, considers only the blind-spot loss $\mathcal{L}_{bsp}$ and establishes a self-supervised $J$-invariant baseline. The second and the third ones are not $J$-invariant, both of them include reconstruction loss and invariance loss calculated before (Eq.~\eqref{eq:deconv}) or after (Eq.~\eqref{eq:reconv}) the PSF convolution. By default, we calculate boundary regularization loss and invariance loss from the same output. We set $\lambda_{inv}$ and $\lambda_{inv\,(d)}$ to $2$~\cite{xie2020noise2same}, $\lambda_{bound}$ and $\lambda_{bound\,(d)}$ to $0.1$~\cite{kobayashi2020image} unless otherwise specified.

\begin{equation}
\label{eq:loss}
\begin{split}
\mathcal{L}(f) = 
& \lambda_{bsp} \, \mathbb{E}_J\mathbb{E}_x \|g(f(\mathbf{x}_{J^c}))_J - \mathbf{x}_J\| ^2 + \\
& \lambda_{rec} \, \mathbb{E}_x \|g(f(\mathbf{x})) - \mathbf{x}\|^2 / m + \\
& \lambda_{inv} \, \mathbb{E}_J \left[ \mathbb{E}_x \|g(f(\mathbf{x}))_J - g(f(\mathbf{x}_{J^c}))_J \|^2 / |J| \right]^{1/2} + \\
& \lambda_{inv\,(d)} \, \mathbb{E}_J \left[ \mathbb{E}_x \|f(\mathbf{x})_J - f(\mathbf{x}_{J^c})_J \|^2 / |J| \right]^{1/2} + \\
& \lambda_{bound} \, \mathbb{E}_x \left( |min - g(f(\mathbf{x}))| + |g(f(\mathbf{x})) - max| \right) + \\
& \lambda_{bound\,(d)} \, \mathbb{E}_x \left( |min - f(\mathbf{x})| + |f(\mathbf{x}) - max| \right)
\end{split}
\end{equation}

\begin{align}
    \mathcal{L}(f)_{\text{Noise2Self}} &= \mathcal{L}_{bsp} \label{eq:bsp} \\
    \mathcal{L}(f)_{\text{Noise2Same}} &= \mathcal{L}_{rec} + \lambda_{inv}\,\mathcal{L}_{inv} + \lambda_{bound}\,\mathcal{L}_{bound} \label{eq:reconv}  \\
    \mathcal{L}(f)_{\text{Noise2Same\,(d)}} &= \mathcal{L}_{rec} + \lambda_{inv\,(d)}\,\mathcal{L}_{inv\,(d)} + \lambda_{bound\,(d)}\,\mathcal{L}_{bound\,(d)} \label{eq:deconv}
\end{align}

\subsubsection{Neural network architecture}
\label{sec:nn}
For 2D data, we use the previously proposed~\cite{krull2019noise2void,xie2020noise2same} variation of a U-Net~\cite{ronneberger2015u} architecture without modifications. The fully-convolutional network of depth 3 consists of an encoder and a decoder with concatenating skip connections at corresponding levels. The first convolutional layer outputs 96 features, and this number doubles with every downsampling step. For 3D data, we modify the network to have 48 output features from the first convolution. We also replace the concatenation operation with an addition in skip connections. Such modification allows for a reduction in training time for 3D convolutional networks without substantial performance sacrifices. We implement the network using PyTorch~\cite{paszke2019pytorch}.

\subsubsection{Training}
\label{sec: training}
During training, we sample a patch ($128 \times 128$ pixels for 2D images, $64 \times 64 \times 64$ for 3D), randomly rotate and flip it. We use batch sizes of $16$ for 2D images and $4$ for 3D images. Each image is standardized individually to zero mean and unit variance. For the masked forward pass, we randomly replace $0.5\%$ of pixels from each training image with Gaussian noise ($\sigma=0.2$)~\cite{xie2020noise2same}. We use Adam~\cite{kingma2017adam} optimizer and multiply its learning rate from $4\cdot10^{-4}$ by $0.5$ every $500$ steps out of 3k total for 2D data~\cite{kobayashi2020image} and every 2k steps out of 15k total for 3D data. We train and evaluate all models on a single NVIDIA V100 16GB GPU.  

\subsubsection{Inference}
\label{sec:inference}
For inference, we do only unmasked forward pass and discard the PSF convolution during inference~\cite{kobayashi2020image}. We use the last model checkpoint since we do not possess any reliable validation metric. For large images that do not fit in GPU memory, we predict by overlapping patches and stitch the output together using pyramid weights~\cite{Khvedchenya_Eugene_2019_PyTorch_Toolbelt} which allows for avoiding edge artifacts. In 3D data, we predict patches of $128 \times 128 \times 128$ with overlaps of $32$ pixels.

\subsubsection{Fast Fourier Transform Convolution}
\label{sec:fft}
We use FFT for the convolution with a fixed PSF during training. Convolution operation $x \ast k$ is equivalent to element-wise matrix multiplication in Fourier space $F^{-1}(F(x) \odot F(k))$. If image $x$ is $M \times M$ pixels and kernel $k$ is $N \times N$ and both of them have dimensionality of $d$, the complexity of an ordinary convolution is $O(M^d N^d)$, while the complexity of a Fourier convolution is $O(M^d d\log M)$ and does not depend on $N$. With large enough $N$, convolution in Fourier space becomes computationally cheaper, and large kernels (even of the size of the image~\cite{sage2017deconvolutionlab2}) are common for PSF. It is recommended to use FFT with 2D kernels for $N > 25$ and with 3D kernels for $N > 9$~\cite{Frank_Odom_fkodom_fft-conv-pytorch}. 

\section{Experiments}
\label{sec: experiments}

We validate our approach on 2D and 3D data in a single image deconvolution task. In all the experiments we train one model per image to obtain a lower bound for the extreme self-supervised case. Every image is normalized within $[0, 1]$. We generate blurry images by convolving them with a realistic Richards \& Wolf PSF of a $0.8\text{NA}$ 16x microscope objective with $17 \times 17$ or $17 \times 17 \times 17$ $0.406$-micron pixels. Then we add a mixture of Poisson ($\alpha=0.001$), Gaussian ($\sigma = 0.1$), and salt-and-pepper (only for 2D images, $p=0.01$) noise to these images and quantize them with $10$ bit precision~\cite{kobayashi2020image}.

We compare the results against classical Lucy-Richardson (LR) deconvolution algorithm~\cite{richardson1972bayesian} and deep learning Self-Supervised Inversion (SSI)~\cite{kobayashi2020image} algorithm. For LR evaluate the results after $2$, $5$, $10$, and $20$ iterations to observe the trade-off between deconvolution quality and noise amplification and report the best ones. We do not include in comparison other classical methods such as Conjugate Gradient optimization~\cite{chambolle2011first} and Chambole-Pock primal-dual inversion~\cite{chambolle2011first}, because they were shown inferior to both LR and SSI~\cite{kobayashi2020image}. We rerun the baseline experiments ourselves wherever possible. 

We evaluate the model performance against clean images using root mean squared error (RMSE), peak signal-to-noise ratio (PSNR)~\cite{yuanji2003image}, and structural similarity index(SSIM)~\cite{wang2003multiscale}. For PSNR and SSIM, we explicitly set the data range to the true values $[0, 1]$. Since it was not done in prior work~\cite{kobayashi2020image}, some values in Table~\ref{tab:ssi-results} are missing to avoid confusion. Additionally, for 2D images, we report mutual information (MI)~\cite{russakoff2004image} and spectral mutual information (SMI)~\cite{kobayashi2020image}. For all metrics except RMSE, higher is better.

\begin{figure}[h]
\begin{center}
\includegraphics[width=0.99\linewidth]{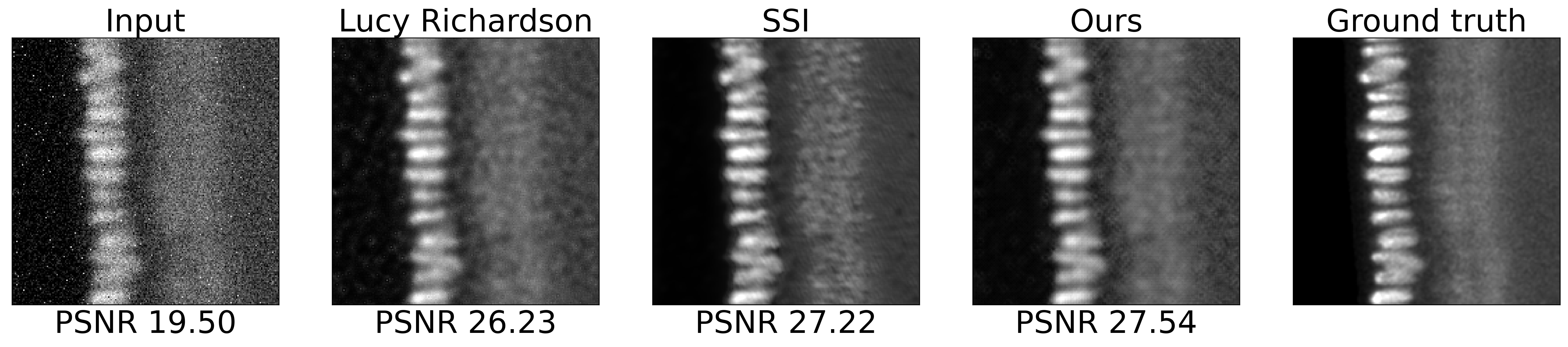}
\includegraphics[width=0.99\linewidth,trim={0 0 0 1.2cm},clip]{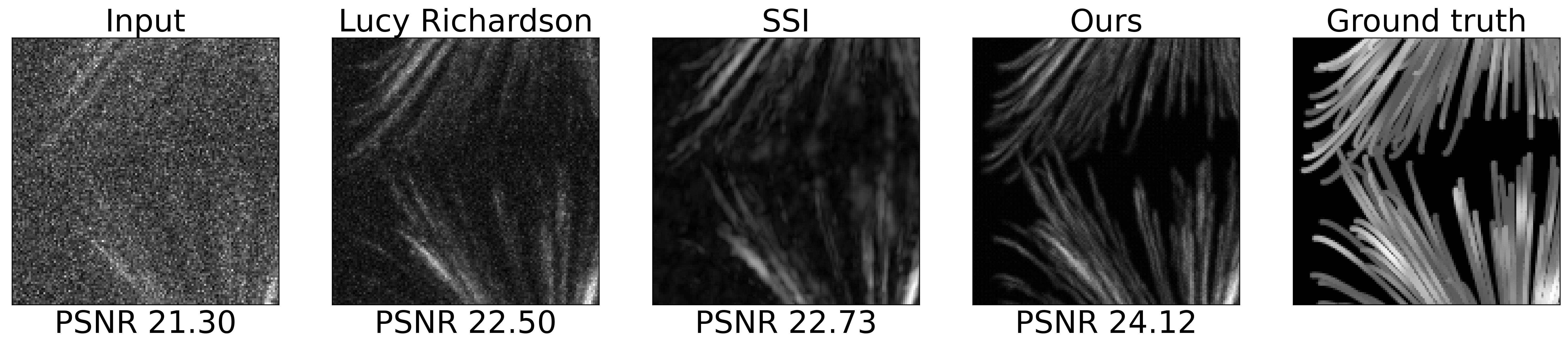}
\end{center}
   \caption{Example deconvolution results on \emph{Drosophila} 2D image (top), microtubules 3D image, frontal max projection (bottom), magnified. For Lucy Richardson, the result is shown after the best iteration. Our algorithm uses loss from Eq.~\eqref{eq:deconv} with $\lambda_{inv\,(d)}=2$, $\lambda_{bound\,(d)}=0.1$. PSNR metric is reported for the presented images.}
\label{fig:ssi}
\end{figure}

\subsection{2D dataset}

First, we tested the deconvolution framework performance for the 2D case on a benchmark dataset of 22 single-channeled images~\cite{kobayashi2020image} with size varying from $512 \times 512$ to $2592 \times 1728$ pixels. We compared three different variants of loss function, Eqs.~\eqref{eq:bsp}--~\eqref{eq:deconv} (Table~\ref{tab:ssi-results-loss}). Loss~\eqref{eq:deconv} performed best with $\lambda_{inv\,(d)} = 2$ and $\lambda_{bound\,(d)} = 0.1$, achieving $\text{PSNR}=22.79$, $\text{SSIM}=0.46$, and $\text{RMSE}=0.078$.

\begin{table}[h]
\centering
\caption{Denoising results for the 2D dataset with different coefficients for loss~\eqref{eq:loss} components: $\lambda_{bsp}$~--- coefficient for the blind-spot loss, $\lambda_{rec}$~--- coefficient for the reconstruction loss, $\lambda_{inv\,(d)}$~--- coefficient for deconvolved invariance loss, $\lambda_{inv}$~--- coefficient for reconvolved invariance loss, $\lambda_{bound}$ and $\lambda_{bound\,(d)}$~--- coefficients for boundary regularization loss for reconvolved and deconvolved images. For all metrics except RMSE, higher is better. The best values are highlighted in bold.}
\label{tab:ssi-results-loss}
\begin{tabular}{lccccccccc}
\toprule
 & $\lambda_{bsp}$ & $\lambda_{rec}$ & $\lambda_{inv\,(d)}$ & $\lambda_{inv}$ & $\lambda_{bound\,(d)}$ & $\lambda_{bound}$ & PSNR $\uparrow$  & SSIM $\uparrow$ & RMSE $\downarrow$   \\
\midrule
input        &   &   &   &   &     &     & 17.77 &  0.18 & 0.130 \\
\midrule
$\mathcal{L}(f)_{\text{Noise2Self}}$     & 1 & 0 & 0 & 0 & 0   & 0   & 22.49 &  0.27 & 0.079 \\
 & 1 & 0 & 0 & 0 & 0   & 0.1 & 22.51 &  0.43 & 0.079 \\ 
\midrule
$\mathcal{L}(f)_{\text{Noise2Same}}$     & 0 & 1 & 0 & 2 & 0   & 0   & 21.49 &  0.23 & 0.089 \\
 & 0 & 1 & 0 & 2 & 0   & 0.1 & 21.81 &  0.41 & 0.085 \\
\midrule
$\mathcal{L}(f)_{\text{Noise2Same (d)}}$     & 0 & 1 & 2 & 0 & 0   & 0   & \textbf{22.79} & 0.46 &  \textbf{0.078} \\
& 0 & 1 & 2 & 0 & 0.1 & 0 & 22.67 &  \textbf{0.46} & 0.078   \\ 
\bottomrule                 
\end{tabular}
\end{table}

We compared our model against the baseline LR and SSI methods (Table~\ref{tab:ssi-results}, Figure~\ref{fig:ssi}). It performed best by PSNR, SMI, and RMSE, and showed similar results by SSIM ($0.01$ less than LR) and MI ($0.01$ less than SSI).

\begin{table}[h]
\centering
\caption{Our best-performing method for 2D data compared to Lucy Richardson (LR)~\cite{richardson1972bayesian} and Self-Supervised Inversion (SSI)~\cite{kobayashi2020image} baselines by denoising metrics (see Section~\ref{sec: experiments} for details), training and inference time (measured for \emph{Drosophila} image $1352 \times 532$ pixels). For all metrics except RMSE, higher is better. The best values are highlighted in bold. Our algorithm uses loss~\eqref{eq:deconv} with $\lambda_{inv\,(d)}=2$, $\lambda_{bound\,(d)}=0.1$}
\label{tab:ssi-results}
\begin{tabular}{lccccccc} 
\toprule
& PSNR $\uparrow$  & SSIM $\uparrow$  &  MI $\uparrow$    &   SMI $\uparrow$ & RMSE $\downarrow$   & Train $t$ (s) & Inference $t$ (ms)                           \\ 
\midrule
input          & 17.8  & 0.18 & 0.07 & 0.18 & 0.131 & -   & - \\ \midrule
LR $n = 2$               & 22.0  & \textbf{0.47} & 0.13 & 0.25 & 0.086 & -   & 46 \\
LR $n = 5$               & 22.2  & 0.44 & 0.12 & 0.25 & 0.082 & -   & 48 \\
SSI (reported~\cite{kobayashi2020image})         & 22.5  &  -   & \textbf{0.14} & 0.27 & -     & 238 & 26 \\
SSI (reproduced)        & 22.0  &  0.46   & 0.14 & 0.26 & -     & 442 & 26 \\
ours                     & \textbf{22.8}  & 0.46 & 0.13 & \textbf{0.30} & \textbf{0.078} & 856 & 53 \\
\bottomrule
\end{tabular}
\end{table}


\subsection{3D dataset}



We evaluated the framework performance for the 3D case on a single synthetic image of microtubules~\cite{sage2017deconvolutionlab2} of size $128 \times 256 \times 512$. From the three variants of loss function, Eqs.~\eqref{eq:bsp}--~\eqref{eq:deconv}, loss~\eqref{eq:reconv} performed best with $\lambda_{inv} = 2$ and $\lambda_{bound} = 0$ showing the results $\text{PSNR}=24.08$, $\text{SSIM}=0.39$, and $\text{RMSE}=0.063$ (Table~\ref{tab:mt-results-loss}).

\begin{table}[h]
\centering
\caption{Denoising results for the 3D dataset with different coefficients for loss~\eqref{eq:loss} components: $\lambda_{bsp}$~--- coefficient for the blind-spot loss, $\lambda_{rec}$~--- coefficient for the reconstruction loss, $\lambda_{inv\,(d)}$~--- coefficient for deconvolved invariance loss, $\lambda_{inv}$~--- coefficient for reconvolved invariance loss, $\lambda_{bound}$ and $\lambda_{bound\,(d)}$~--- coefficients for boundary regularization loss for reconvolved and deconvolved images. For all metrics except RMSE, higher is better. The best values are highlighted in bold.}
\label{tab:mt-results-loss}
\begin{tabular}{lccccccccc}
\toprule
 & $\lambda_{bsp}$ & $\lambda_{rec}$ & $\lambda_{inv\,(d)}$ & $\lambda_{inv}$ & $\lambda_{bound\,(d)}$ & $\lambda_{bound}$ & PSNR $\uparrow$  & SSIM $\uparrow$ & RMSE $\downarrow$   \\
\midrule
input        &   &   &   &   &     &     & 21.21 & 0.64   & 0.087 \\
\midrule

$\mathcal{L}(f)_{\text{Noise2Self}}$   & 1 & 0 & 0 & 0 & 0   & 0   & 23.71 & 0.39 & 0.065 \\
                                       & 1 & 0 & 0 & 0 & 0   & 0.1 & 23.71 & 0.39 & 0.065        \\ 
\midrule
$\mathcal{L}(f)_{\text{Noise2Same}}$   & 0 & 1 & 0 & 2 & 0   & 0   & \textbf{24.08} & 0.39 & \textbf{0.063} \\
                                       & 0 & 1 & 0 & 2 & 0   & 0.1 & 24.02 & 0.39 & 0.063  \\
\midrule
$\mathcal{L}(f)_{\text{Noise2Same (d)}}$   & 0 & 1 & 2 & 0 & 0   & 0 & 21.05 & \textbf{0.82} & 0.089 \\
                                           & 0 & 1 & 2 & 0 & 0.1 & 0 & 21.05 & \textbf{0.82} & 0.089 \\ 
   
\bottomrule     


\end{tabular}
\end{table}

We also compared the best-performing model against the LR and SSI methods (Table~\ref{tab:mt-results}, Figure~\ref{fig:ssi}). Our model surpassed baselines by a substantial margin ($+1.5$ PSNR against SSI and $+1.7$ against LR).

\begin{table}[h]
\centering
\caption{Our best performing method for 3D data compared to Lucy Richardson (LR)~\cite{richardson1972bayesian} and Self-Supervised Inversion (SSI)~\cite{kobayashi2020image} baselines by denoising metrics (see Section~\ref{sec: experiments} for details), training and inference time. For all metrics except RMSE, higher is better. The best values are highlighted in bold. Our algorithm uses loss~\eqref{eq:reconv} with $\lambda_{inv}=2$, $\lambda_{bound}=0$. Training time is reported with and without FFT (in brackets)}
\label{tab:mt-results}
\begin{tabular}{lccccccc} 
\toprule
& PSNR $\uparrow$  & SSIM $\uparrow$  & RMSE $\downarrow$   & Train $t$ (s) & Inference $t$ (ms)                           \\ 
\midrule
input  & 21.2  & 0.64 & 0.087 & - & - \\ \midrule
LR $n = 2$       & 22.4  & 0.32 & 0.076 & - & 69 \\
SSI              & 22.6  & 0.39 & 0.074 & 2814 (6580) & 578 \\
ours             & \textbf{24.1}  & \textbf{0.39} & \textbf{0.062} & 8160 (18900) & 734 \\
\bottomrule
\end{tabular}
\end{table}

\section{Discussion}
\label{sec: discussion}

Our Siamese neural network was superior in single-image deconvolution against both classical and deep learning baselines. The architecture is simple and did not require additional tricks such as masking schedule or adding noise to gradients~\cite{kobayashi2020image} for training. Training converged to similar performance for all of the several tested random weight initializations.

Loss function performance was inconsistent between the 2D and 3D cases. Invariance loss applied to deconvolved representation proved best for 2D data but failed to give an advantage in 3D: despite the high SSIM, images appear noisy. We hypothesize that the optimal solution depends on the data; \eg in microtubules, the 3D dataset signal is very sparse. Boundary loss did not drastically affect the training of Siamese networks.

Despite performance superiority, Siamese networks are twice more expensive in computation because they require two forward passes through the neural network. Additionally, Noise2Same U-Net architecture has 10x more trainable parameters than SSI (5.75M against 0.55M in 2D). This problem is exacerbated by the necessity of convolutions with large PSF kernels which are not optimized in modern GPUs. We propose to alleviate this problem by using FFT for convolution with PSF. It leads to 2.3x speed improvement for 3D deconvolution with PSF of size $17\times17\times17$, and this advantage will grow for larger kernels. For example, while training with a PSF of $31\times31\times31$ we observed a 14x speedup.

\section{Conclusion}
\label{sec:conclusion}

Blind-spot networks~\cite{krull2019noise2void,batson2019noise2self} were seminal in self-supervised denoising and performed comparably to supervised methods~\cite{lehtinen2018noise2noise,weigert2018content}. Their success was later translated to image deconvolution~\cite{kobayashi2020image}, a more complicated inverse problem. However, it was shown that $J$-invariance leads to suboptimal performance in denoising because masked pixels contain useful bits of information~\cite{xie2020noise2same}. 
In this work, we presented a novel unified image deconvolution framework, which generalized the accumulated prior advances, and set a new standard for image quality enhancement performance. We investigated the contributions of various self-supervised loss components and empirically identified the optimal usage scenarios for 2D and 3D data. We also proposed using Fast Fourier Transform for the training of deconvolution neural networks, which drastically speeds up the computation, especially for large kernels. This advantage allows one to use our method for non-confocal microscopy with extremely big point spread functions.   

\section*{Acknowledgements}

This work was funded by Revvity Inc. (VLTAT19682) and Wellcome Trust (206194). We thank High Performance Computing Center of the Institute of Computer Science at the University of Tartu for the provided computing power.

%
%
%
\bibliographystyle{splncs04}
\bibliography{ref}

\clearpage
\include{supplementary}

\end{document}

%% file: supplementary.tex
\renewcommand{\thetable}{S\arabic{table}}
\renewcommand{\thefigure}{S\arabic{figure}}
\setcounter{figure}{0}
\setcounter{table}{0}

\section*{Supplementary material}

\begin{figure}[h]
\begin{center}
\includegraphics[width=0.99\linewidth]{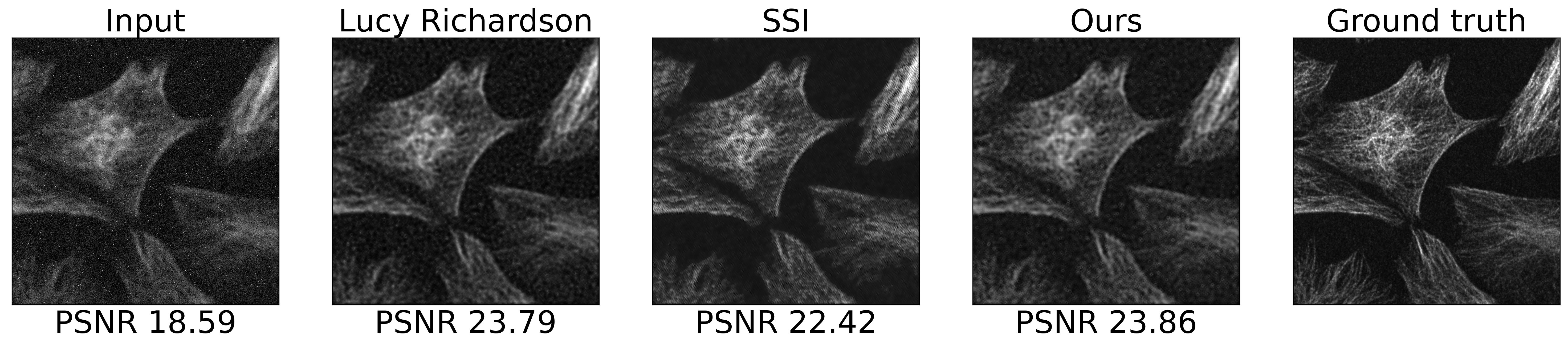}
\end{center}
   \caption{Example deconvolution results on \emph{Tubules} 2D image, magnified. For Lucy Richardson the result is shown after 5 iterations. Our algorithm uses loss from Eq.~\eqref{eq:deconv} with $\lambda_{inv\,(d)}=2$, $\lambda_{bound\,(d)}=0.1$. PSNR metric is reported for the presented image.}
\label{fig:ssi:supp}
\end{figure}

\begin{table}[h]
\centering
\caption{Our best-performing method for 2D data compared to Lucy Richardson (LR) baseline by denoising metrics (see Section~\ref{sec: experiments} for details), training and inference time (measured for \emph{Drosophila} image $1352 \times 532$ pixels). For all metrics except RMSE, higher is better. The best values are highlighted in bold. Our algorithm uses loss~\eqref{eq:deconv} with $\lambda_{inv\,(d)}=2$, $\lambda_{bound\,(d)}=0.1$}
\label{tab:ssi-results:supp}
\begin{tabular}{lccccccc} 
\toprule
& PSNR $\uparrow$  & SSIM $\uparrow$  &  MI $\uparrow$    &   SMI $\uparrow$ & RMSE $\downarrow$   & Train $t$ (s) & Inference $t$ (ms)                           \\ 
\midrule
input          & 17.8  & 0.18 & 0.07 & 0.18 & 0.131 & -   & - \\ \midrule
LR $n = 2$               & 22.0  & \textbf{0.47} & 0.13 & 0.25 & 0.086 & -   & 46 \\
LR $n = 5$               & 22.2  & 0.44 & 0.12 & 0.25 & 0.082 & -   & 48 \\
LR $n = 10$              & 21.1  & 0.36 & 0.10 & 0.25 & 0.091 & -   & 62 \\
LR $n = 20$              & 18.5  & 0.26 & 0.08 & 0.19 & 0.119 & -   & 97 \\
ours                     & \textbf{22.8}  & 0.46 & \textbf{0.13} & \textbf{0.30} & \textbf{0.078} & 856 & 53 \\
\bottomrule
\end{tabular}
\end{table}

\begin{figure}[h]
\begin{center}
\includegraphics[width=0.99\linewidth]{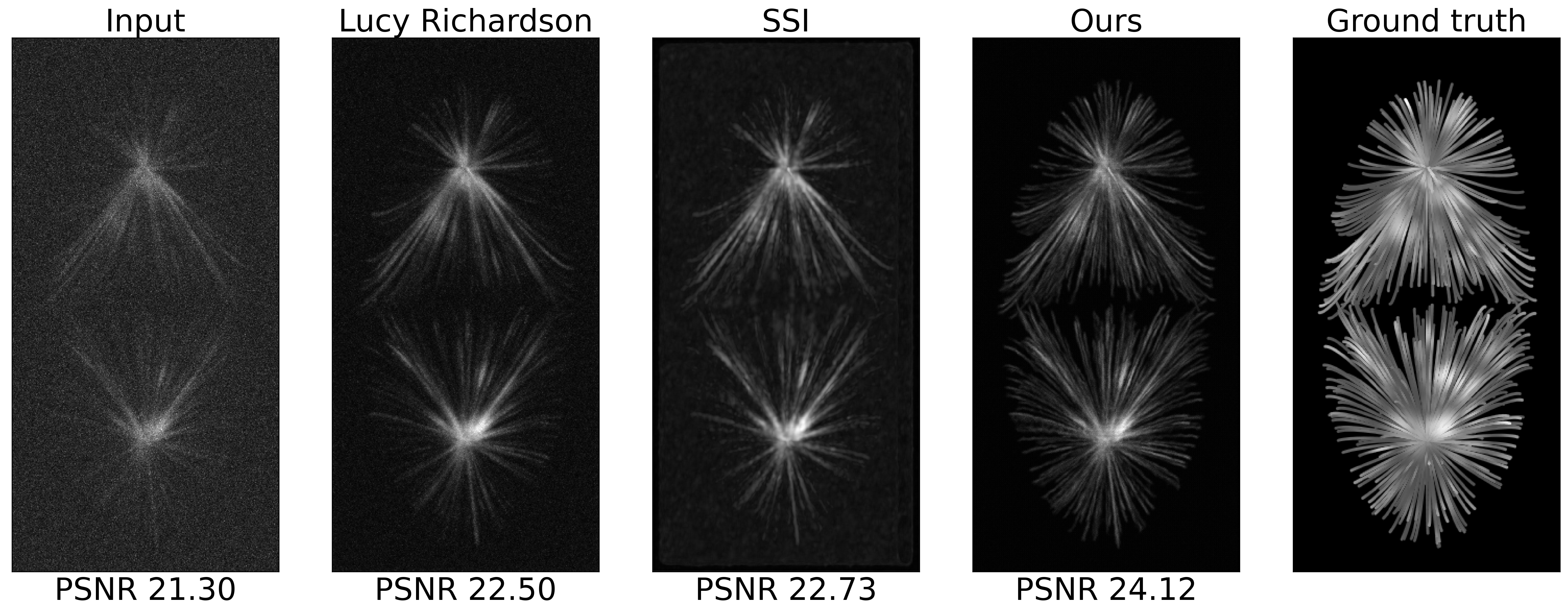}
\includegraphics[width=0.99\linewidth]{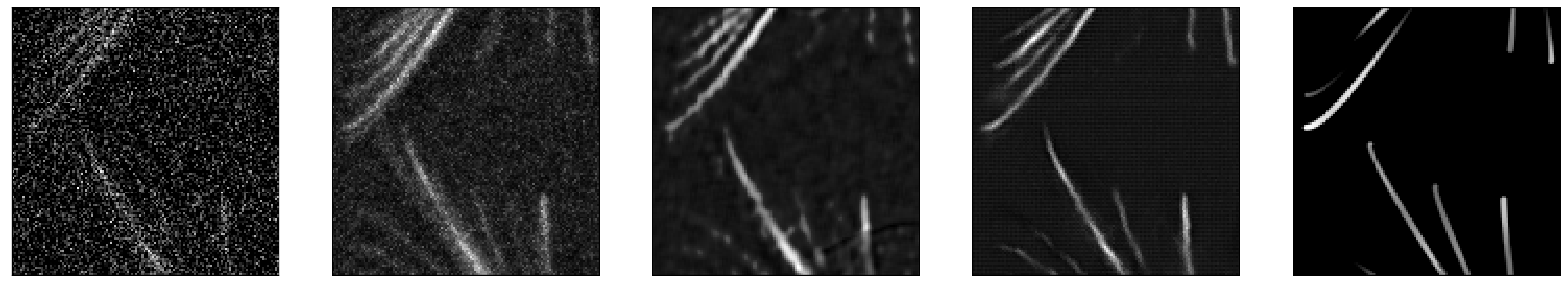}
\includegraphics[width=0.99\linewidth]{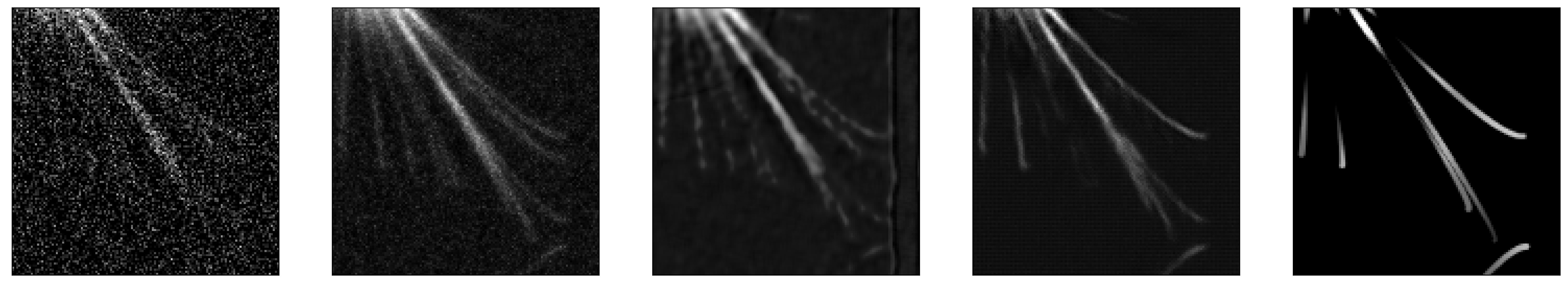}
\end{center}
   \caption{Example deconvolution results on microtubules 3D image, frontal max projection (top), cross-section at plane 68 in two magnified crops (middle, bottom). For Lucy Richardson the best result is shown after 2 iterations. Our algorithm uses loss from Eq.~\eqref{eq:reconv} with $\lambda_{inv}=2$, $\lambda_{bound}=0$.}
\label{fig:mt:supp}
\end{figure}

\begin{table}[h]
\centering
\caption{Our best-performing method for 3D data compared to Lucy Richardson (LR) baseline by denoising metrics (see Section~\ref{sec: experiments} for details), training, and inference time. For all metrics except RMSE, higher is better. The best values are highlighted in bold. Our algorithm uses loss~\eqref{eq:reconv} with $\lambda_{inv}=2$, $\lambda_{bound}=0.1$. Training time is reported with and without FFT (in brackets)}
\label{tab:mt-results:supp}
\begin{tabular}{lccccccc} 
\toprule
& PSNR $\uparrow$  & SSIM $\uparrow$  & RMSE $\downarrow$   & Train $t$ (s) & Inference $t$ (ms)                           \\ 
\midrule

input  & 21.2  & 0.64 & 0.087 & - & - \\ \midrule
LR $n = 2$       & 22.4  & 0.32 & 0.076 & - & 69 \\
LR $n = 5$       & 21.8  & 0.33 & 0.080 & - & 167 \\
LR $n = 10$      & 21.5  & 0.44 & 0.084 & - & 222 \\
LR $n = 20$      & 21.3  & \textbf{0.55} & 0.086 & - & 369 \\
ours             & \textbf{24.1}  & 0.39 & \textbf{0.062} & 8160 (18900) & 734 \\
\bottomrule
\end{tabular}
\end{table}
